\pdfoutput=1

\documentclass[11pt]{article}

\usepackage{naacl2021}

\usepackage{times}
\usepackage{latexsym}

\usepackage[T1]{fontenc}

\usepackage[utf8]{inputenc}

\usepackage{microtype}

\usepackage{amsmath}
\usepackage{amssymb}
\usepackage{mathrsfs}
\usepackage{xcolor}
\usepackage{amsthm}
\usepackage{amsfonts}
\usepackage{graphicx}
\usepackage{stmaryrd}
\usepackage{algorithm}
\usepackage{xr}

\usepackage{url}
\usepackage{array}
\usepackage{booktabs}
\usepackage{balance}
\usepackage{colortbl}
\usepackage{makecell}
\usepackage{soul}
\usepackage{blindtext}
\usepackage{multirow}
\usepackage{subcaption}
\usepackage{amssymb}
\usepackage{pifont}
\newcommand{\cmark}{\ding{51}}%
\newcommand{\xmark}{\ding{55}}%

\newcolumntype{P}[1]{>{\centering\arraybackslash}p{#1}}

\usepackage{graphicx,wrapfig,lipsum}

\usepackage{amsmath,amsfonts,bm}

\def\eqref#1{(\ref{#1})}

\def\1{\bm{1}}

\def\vw{{\bm{w}}}

\DeclareMathAlphabet{\mathsfit}{\encodingdefault}{\sfdefault}{m}{sl}
\SetMathAlphabet{\mathsfit}{bold}{\encodingdefault}{\sfdefault}{bx}{n}

\usepackage{commath}
\usepackage{amssymb}
\usepackage{amsmath,amsfonts,bm}

\newcommand{\dataname}{SVAMP}

\title{Are NLP Models really able to Solve Simple Math Word Problems?}

\author{Arkil Patel \quad Satwik Bhattamishra \quad Navin Goyal \\
		Microsoft Research India \\
		\texttt{arkil.patel@gmail.com, \{t-satbh,navingo\}@microsoft.com}}

\begin{document}
\maketitle
\begin{abstract}
The problem of designing NLP solvers for math word problems (MWP) has seen sustained research activity and steady gains in the test accuracy. Since existing solvers achieve high performance on the benchmark datasets for elementary level MWPs containing one-unknown arithmetic word problems, such problems are often considered ``solved'' with the bulk of research attention moving to more complex MWPs. In this paper, we restrict our attention to English MWPs taught in grades four and lower. We provide strong evidence that the existing MWP solvers rely on shallow heuristics to achieve high performance on the benchmark datasets. To this end, we show that MWP solvers that do not have access to the question asked in the MWP can still solve a large fraction of MWPs. Similarly, models that treat MWPs as bag-of-words can also achieve surprisingly high accuracy. Further, we introduce a challenge dataset, SVAMP, created by applying carefully chosen variations over examples sampled from existing datasets. The best accuracy achieved by state-of-the-art models is substantially lower on SVAMP, thus showing that much remains to be done even for the simplest of the MWPs. 
\end{abstract}

\section{Introduction}
 
A Math Word Problem (MWP) consists of a short natural language narrative describing a state of the world and poses a question about some unknown quantities (see Table~\ref{tab:intro_ex} for some examples). MWPs are taught in primary and higher schools.
The MWP task is a type of semantic parsing task where given an MWP the goal is to generate an expression (more generally, equations), which can then be evaluated to get the answer.
The task is challenging because a machine needs to extract relevant information from  natural language text as well as perform mathematical reasoning to solve it. The complexity of MWPs can be measured along multiple axes, 
e.g., reasoning and linguistic complexity and world and domain knowledge. A combined complexity measure is the \emph{grade level} of an MWP, which is the grade in which similar MWPs are taught. Over the past few decades many approaches have been developed to solve MWPs with significant activity in the last decade \cite{8703135}. 
\definecolor{azure}{rgb}{0.0, 0.5, 1.0}
\definecolor{amber}{rgb}{1.0, 0.49, 0.0}
\definecolor{americanrose}{rgb}{1.0, 0.01, 0.24}
\definecolor{amethyst}{rgb}{0.6, 0.4, 0.8}
\definecolor{applegreen}{rgb}{0.55, 0.61, 0.0}
\definecolor{ballblue}{rgb}{0.13, 0.67, 0.8}
\definecolor{bazaar}{rgb}{0.6, 0.47, 0.48}
\begin{table}[t]
	\small{\centering
		\begin{tabular}{m{18em}m{3em}}
			\toprule
			\textsc{\textcolor{green!45!blue}{Problem:}}& \\
			\multicolumn{2}{p{23em}}{\textcolor{green!50!red}{Text:} Jack had 8 pens and Mary had 5 pens. Jack gave 3 pens to Mary. How many pens does Jack have now?} \\
			\textcolor{green!50!red}{Equation:} 8 - 3 = 5 \\
			\midrule
			\textsc{\textcolor{green!45!blue}{Question Sensitivity Variation:}}& \\
			\multicolumn{2}{p{23em}}{\textcolor{green!50!red}{Text:} Jack had 8 pens and Mary had 5 pens. Jack gave 3 pens to Mary. How many pens does \textcolor{red!65!blue}{Mary} have now?} \\
			\textcolor{green!50!red}{Equation:} 5 + 3 = 8 \\
			\midrule
			\textsc{\textcolor{green!45!blue}{Reasoning Ability Variation:}}& \\
			\multicolumn{2}{p{23em}}{\textcolor{green!50!red}{Text:} Jack had 8 pens and Mary had 5 pens. \textcolor{red!65!blue}{Mary} gave 3 pens to \textcolor{orange!25!brown}{Jack}. How many pens does Jack have now?} \\
			\textcolor{green!50!red}{Equation:} 8 + 3 = 11 \\
			\midrule
			\textsc{\textcolor{green!45!blue}{Structural Invariance Variation:}}& \\
			\multicolumn{2}{p{23em}}{\textcolor{green!50!red}{Text:} \textcolor{red!65!blue}{Jack gave 3 pens} to Mary. If \textcolor{orange!25!brown}{Jack had 8 pens} and Mary had 5 pens initially, how many pens does Jack have now?} \\
			\textcolor{green!50!red}{Equation:} 8 - 3 = 5 \\
			\bottomrule
		\end{tabular}
		\caption{\label{tab:intro_ex} Example of a Math Word Problem along with the types of variations that we make to create \dataname{}.}
	}
\end{table}

MWPs come in many varieties. Among the simplest are the one-unknown arithmetic word problems where the output is a mathematical expression involving numbers and one or more arithmetic operators ($+, -, *, /$). Problems in Tables~\ref{tab:intro_ex} and \ref{tab:attn_wts} are of this type. More complex MWPs may have systems of equations as output or involve other operators or may involve more advanced topics and specialized knowledge.
Recently, researchers have started focusing on solving such MWPs, e.g. multiple-unknown linear word problems \cite{huang-etal-2016-well}, geometry \cite{sachan-xing-2017-learning} and probability \cite{amini-etal-2019-mathqa}, believing that existing work can handle one-unknown arithmetic MWPs well \cite{qin2020semanticallyaligned}. In this paper, we question the capabilities of the state-of-the-art (SOTA) methods to robustly solve even the simplest of MWPs suggesting that the above belief is not well-founded.

In this paper, we provide concrete evidence to show that existing methods use shallow heuristics to solve a majority of word problems in the benchmark datasets. We find that existing models are able to achieve reasonably high accuracy on MWPs from which the question text has been removed leaving only the narrative describing the state of the world. This indicates that the models can rely on superficial patterns present in the narrative of the MWP and 
achieve high accuracy without even looking at the question. In addition, we show that a model without word-order information (i.e., the model treats the MWP as a bag-of-words) can also solve the majority of MWPs in benchmark datasets. 

%

The presence of these issues in existing benchmarks makes them unreliable for measuring the performance of models. Hence, we create a challenge set called \dataname{} (\textbf{S}imple \textbf{V}ariations on \textbf{A}rithmetic \textbf{M}ath word \textbf{P}roblems; pronounced \textit{swamp}) of one-unknown arithmetic word problems with grade level up to 4 by applying simple variations over word problems in an existing dataset (see Table \ref{tab:intro_ex} for some examples). \dataname{} further highlights the brittle nature of existing models when trained on these benchmark datasets. On evaluating SOTA models on \dataname{}, we find that they are not even able to solve half the problems in the dataset. This failure of SOTA models on \dataname{} points to the extent to which they rely on simple heuristics in training data to make their prediction.



Below, we summarize the two broad contributions of our paper.

\begin{itemize}
	\item We show that the majority of problems in benchmark datasets can be solved by shallow heuristics lacking word-order information or lacking question text.
	\item We create a challenge set called \dataname{} \footnote{The dataset and code are available at: \href{https://github.com/arkilpatel/SVAMP}{https://github.com/arkilpatel/SVAMP}} for more robust evaluation of methods developed to solve elementary level math word problems.
\end{itemize}

\section{Related Work}\label{sec:relwork}

\textbf{Math Word Problems.} A wide variety of methods and datasets have been proposed to solve MWPs; e.g. statistical machine learning \cite{roy-roth-tacl}, 
semantic parsing \cite{huang-etal-2017-learning-fine} and most recently deep learning \cite{wang-etal-2017-deep-neural, gts-ijcai-2019, zhang-etal-2020-graph};
see \cite{8703135} for an extensive survey. 
Many papers have pointed out various deficiencies with previous datasets and proposed new ones to address them. \citet{koncel-kedziorski-etal-2016-mawps} curated the MAWPS dataset from previous datasets 
which along with Math23k \cite{wang-etal-2017-deep-neural} has been used as benchmark in recent works. Recently, ASDiv \cite{miao-etal-2020-diverse} has
been proposed to provide more diverse problems with annotations for equation, problem type and grade level. 
HMWP \cite{qin2020semanticallyaligned} is another newly proposed dataset of Chinese MWPs that includes examples with muliple-unknown variables and requiring non-linear equations to solve them.

\textbf{Identifying artifacts in datasets} has been done for the Natural Language Inference (NLI) task by \citet{mccoy-etal-2019-right}, \citet{poliak-etal-2018-hypothesis}, and \citet{gururangan-etal-2018-annotation}. \citet{rosenman-etal-2020-exposing} identified shallow heuristics in a Relation Extraction dataset. \citet{cai-etal-2017-pay} showed that biases prevalent in the ROC stories cloze task allowed models to yield state-of-the-art results when trained only on the endings. To the best of our knowledge, this kind of analysis has not been done on any Math Word Problem dataset.

\textbf{Challenge Sets} for NLP tasks have been proposed most notably for NLI and machine translation \cite{Belinkov_Glass_Survey, anli, checklist}. 
\citet{gardner2020evaluating} suggested creating \textit{contrast sets} by manually perturbing test instances in small yet meaningful ways that change the gold label. 
We believe that we are the first to introduce a challenge set targeted specifically for robust evaluation of Math Word Problems.

\section{Background}

\subsection{Problem Formulation}\label{formulation}

We denote a Math Word Problem $P$ by a sequence of $n$ tokens $P = (\vw_{1},\ldots, \vw_n)$ where each token $\vw_{i}$ can be either a word from a natural language or a numerical value. The word problem $P$ can be broken down into \emph{body} $B = (\vw_{1},\ldots, \vw_k)$ and \emph{question} $Q = (\vw_{k+1},\ldots, \vw_n)$. The goal is to map $P$ to a valid mathematical expression $E_P$ composed of numbers from $P$ and mathematical operators from the set $\{+,-,/,*\}$ (e.g. $3+5-4$). 
The metric used to evaluate models on the MWP task is Execution Accuracy, which is obtained from comparing the predicted answer (calculated by evaluating $E_P$) with the annotated answer. In this work, we focus only on one-unknown arithmetic word problems. 


\subsection{Datasets and Methods}\label{data_methods}

\begin{table}[t]
	\small{\centering
		\begin{tabular}{p{13em}P{3.5em}P{4em}}
			\toprule
			\textbf{Model} & \textbf{MAWPS}& \textbf{ASDiv-A}\\
			\midrule
			Seq2Seq (S)  & 79.7 & 55.5 \\
			Seq2Seq (R) & 86.7 & 76.9 \\
			\midrule
			GTS (S) \scriptsize{\cite{gts-ijcai-2019}} & 82.6 & 71.4\\
			GTS (R) & 88.5 & 81.2 \\
			\midrule
			Graph2Tree (S) \scriptsize{\cite{zhang-etal-2020-graph}} & 83.7 & 77.4\\
			Graph2Tree (R) & \textbf{88.7} & \textbf{82.2} \\
			\midrule
			Majority Template Baseline\footnotemark{} & 17.7 & 21.2 \\
			\bottomrule
		\end{tabular}
		\caption{\label{tab:std_scores}5-fold cross-validation accuracies ($\uparrow$) of baseline models on datasets. (R) means that the model is provided with RoBERTa pretrained embeddings while (S) means that the model is trained from scratch.}
	}
\end{table}
\footnotetext{Majority Template Baseline is the accuracy when the model always predicts the most frequent \emph{Equation Template}. Equation Templates are explained in Section \ref{data_props}}

Many of the existing datasets are not suitable for our analysis as either they are in Chinese, e.g. Math23k \cite{wang-etal-2017-deep-neural} and HMWP \cite{qin2020semanticallyaligned}, or have harder problem types, e.g. Dolphin18K \cite{Dolphin}. We consider the widely used benchmark MAWPS \cite{koncel-kedziorski-etal-2016-mawps} composed of 2373 MWPs and the arithmetic subset of ASDiv \cite{miao-etal-2020-diverse} called ASDiv-A which has 1218 MWPs mostly up to grade level 4 (MAWPS does not have grade level information). Both MAWPS and ASDiv-A are evaluated on 5-fold cross-validation based on pre-assigned splits.

\noindent We consider three models in our experiments: 

\noindent	\textbf{(a)} \textbf{Seq2Seq} consists of a Bidirectional LSTM Encoder to encode the input sequence and an LSTM decoder with attention \cite{luong-etal-2015-effective} to generate the equation.

	
\noindent	\textbf{(c)} \textbf{GTS} \cite{gts-ijcai-2019} uses an LSTM Encoder to encode the input sequence and a tree-based Decoder to generate the equation. 
	
\noindent	\textbf{(d)} \textbf{Graph2Tree} \cite{zhang-etal-2020-graph} combines a Graph-based Encoder with a Tree-based Decoder.

The performance of these models on both datasets is shown in Table \ref{tab:std_scores}. We either provide RoBERTa \cite{liu2019roberta} pre-trained embeddings to the models or train them from scratch. Graph2Tree \cite{zhang-etal-2020-graph} with RoBERTa embeddings achieves the state-of-the-art for both datasets. Note that our implementations achieve a higher score than the previously reported highest score of 78\% on ASDiv-A \cite{miao-etal-2020-diverse} and 83.7\% on MAWPS \cite{zhang-etal-2020-graph}. The implementation details are provided in Section \ref{implementation} in the Appendix.

\section{Deficiencies in existing datasets}\label{flaws}

Here we describe the experiments that show that there are important deficiencies in MAWPS and ASDiv-A.

\subsection{Evaluation on Question-removed MWPs}

As mentioned in Section \ref{formulation}, each MWP consists of a body $B$, which provides a short narrative on a state of the world and a question $Q$, which inquires about an unknown quantity about the state of the world. For each fold in the provided 5-fold split in MAWPS and ASDiv-A, we keep the train set unchanged while we remove the questions $Q$ from the problems in the test set. Hence, each problem in the test set consists of only the body $B$ without any question $Q$. We evaluate all three models with RoBERTa embeddings on these datasets. The results are provided in Table \ref{tab:prem-only_scores}.

\begin{table}[t]
	\small{\centering
		\begin{tabular}{p{7em}P{4em}P{4em}}
			\toprule
			\textbf{Model} & \textbf{MAWPS}& \textbf{ASDiv-A}\\
			\midrule
			Seq2Seq & 77.4 & 58.7 \\
			GTS & 76.2 & 60.7\\
			Graph2Tree & \textbf{77.7} & \textbf{64.4}\\
			\bottomrule
		\end{tabular}
		\caption{\label{tab:prem-only_scores}5-fold cross-validation accuracies ($\uparrow$) of baseline models on Question-removed datasets.}
	}
\end{table}

The best performing model is able to achieve a 5-fold cross-validation accuracy of 64.4\% on ASDiv-A and 77.7\% on MAWPS. Loosely translated, this means that nearly 64\% of the problems in ASDiv-A and 78\% of the problems in MAWPS can be correctly answered without even looking at the question. This suggests the presence of patterns in the bodies of MWPs in these datasets that have a direct correlation with the output equation.

Some recent works have also demonstrated similar evidence of bias in NLI datasets \cite{gururangan-etal-2018-annotation,poliak-etal-2018-hypothesis}. They observed that NLI models were able to predict the correct label for a large fraction of the standard NLI datasets based on only the hypothesis of the input and without the premise. Our results on question-removed examples of math word problems resembles their observations on NLI datasets and similarly indicates the presence of artifacts that help statistical models predict the correct answer without complete information. Note that even though the two methods appear similar, there is an important distinction. In \citet{gururangan-etal-2018-annotation}, the model is \emph{trained} and tested on hypothesis only examples and hence, the model is forced to find artifacts in the hypothesis during \textit{training}. On the other hand, our setting is more natural since the model is trained in the standard way on examples with both the body and the question. Thus, the model is not explicitly forced to learn based on the body during training and our results not only show the presence of artifacts in the datasets but also suggest that the SOTA models exploit them.

Following \citet{gururangan-etal-2018-annotation}, we attempt to understand the extent to which SOTA models rely on the presence of simple heuristics in the body to predict correctly. We partition the test set into two subsets for each model: problems that the model predicted correctly without the question are labeled \textit{Easy} and the problems that the model could not answer correctly without the question are labeled \textit{Hard}. Table \ref{tab:easy_hard} shows the performance of the models on their respective \textit{Hard} and \textit{Easy} sets. Note that their performance on the full set is already provided in Table \ref{tab:std_scores}. It can be seen clearly that although the models correctly answer many \textit{Hard} problems, the bulk of their success is due to the \textit{Easy} problems. This shows that the ability of SOTA methods to robustly solve word problems is overestimated and that they rely on simple heuristics in the body of the problems to make predictions.

\begin{table}[t]
	\small{\centering
		\begin{tabular}{p{6em}P{2.5em}P{2.5em}P{2.5em}P{2.5em}}
			\toprule
			&\multicolumn{2}{c}{\textbf{MAWPS}} &\multicolumn{2}{c}{\textbf{ASDiv-A}} \\ 
			\cmidrule(lr){2-3}\cmidrule(lr){4-5}
			\textbf{Model} & \textit{Easy}& \textit{Hard}& \textit{Easy}& \textit{Hard}\\
			\midrule
			Seq2Seq & 86.8 & 86.7 & 91.3 & 56.1 \\
			GTS & 92.6 & 71.7 & 91.6 & 65.3  \\
			Graph2Tree & 93.4 & 71.0 & 92.8 & 63.3 \\
			\bottomrule
		\end{tabular}
		\caption{\label{tab:easy_hard}Results of baseline models on the \textit{Easy} and \textit{Hard} test sets.}
	}
\end{table}

\subsection{Performance of a constrained model}

We construct a simple model based on the Seq2Seq architecture by removing the LSTM Encoder and replacing it with a Feed-Forward Network that maps the input embeddings to their hidden representations. The LSTM Decoder is provided with the \textbf{average} of these hidden representations as its initial hidden state. During decoding, an attention mechanism \cite{luong-etal-2015-effective} assigns weights to individual hidden representations of the input tokens. We use either RoBERTa embeddings (non-contextual; taken directly from Embedding Matrix) or train the model from scratch. Clearly, this model does not have access to word-order information. 


Table \ref{tab:simple_scores} shows the performance of this model on MAWPS and ASDiv-A. The constrained model with non-contextual RoBERTa embeddings is able to achieve a cross-validation accuracy of 51.2 on ASDiv-A and an astounding 77.9 on MAWPS. It is surprising to see that a model having no word-order information can solve a majority of word problems in these datasets. These results indicate that it is possible to get a good score on these datasets by simply associating the occurence of specific words in the problems to their corresponding equations. We illustrate this more clearly in the next section.

\begin{table}[t]
	\small{\centering
		\begin{tabular}{p{11em}P{4em}P{4em}}
			\toprule
			\textbf{Model} & \textbf{MAWPS}& \textbf{ASDiv-A}\\
			\midrule
			FFN + LSTM Decoder (S) & 75.1 & 46.3 \\
			FFN + LSTM Decoder (R) & \textbf{77.9} & \textbf{51.2} \\
			\bottomrule
		\end{tabular}
		\caption{\label{tab:simple_scores}5-fold cross-validation accuracies ($\uparrow$) of the constrained model on the datasets. (R) denotes that the model is provided with non-contextual RoBERTa pretrained embeddings while (S) denotes that the model is trained from scratch.}
	}
\end{table}

\subsection{Analyzing the attention weights}

\definecolor{capri}{rgb}{0.0, 0.75, 1.0}
\definecolor{caribbeangreen}{rgb}{0.0, 0.8, 0.6}
\begin{table*}[t]
	\scriptsize{\centering
		\begin{tabular}{p{48.5em}P{8.5em}P{3em}}
			\toprule
			\textbf{Input Problem} & \textbf{Predicted Equation} & \textbf{Answer} \\
			\midrule
			
			
			\scriptsize{John delivered 3 letters at \colorbox{caribbeangreen}{every} house. If he delivered for 8 houses, how many letters did John deliver?} & 3 * 8 & 24 \cmark \\
			\scriptsize{John delivered 3 letters at \colorbox{caribbeangreen}{every} house. He delivered 24 letters in all. How many houses did John visit to deliver letters?} & 3 * 24 & 72 \xmark \\
			\midrule
			\scriptsize{Sam made 8 dollars mowing lawns over the Summer. He charged 2 bucks for \colorbox{caribbeangreen}{each} lawn. How many lawns did he mow?} & 8 / 2 & 4 \cmark \\
			\scriptsize{Sam mowed 4 lawns over the Summer. If he charged 2 bucks for \colorbox{caribbeangreen}{each} lawn, how much did he earn?} & 4 / 2 & 2 \xmark \\
			
			\midrule			
			\scriptsize{10 apples were in the box. 6 are red and the \colorbox{caribbeangreen}{rest} are green. how many green apples are in the box?} & 10 - 6 & 4 \cmark \\
			\scriptsize{10 apples were in the box. \colorbox{caribbeangreen}{Each} apple is either red or green. 6 apples are red. how many green apples are in the box?} & 10 / 6 & 1.67 \xmark \\
			\bottomrule
		\end{tabular}
		\caption{\label{tab:attn_wts} Attention paid to specific words by the constrained model.}
	}
\end{table*}

To get a better understanding of how the constrained model is able to perform so well, we analyze the attention weights that it assigns to the hidden representations of the input tokens. As shown by \citet{wiegreffe-pinter-2019-attention}, analyzing the attention weights of our constrained model is a reliable way to explain its prediction since each hidden representation consists of information about only that token as opposed to the case of an RNN where each hidden representation may have information about the context i.e. its neighboring tokens.

We train the contrained model (with RoBERTa embeddings) on the full ASDiv-A dataset and observe the attention weights it assigns to the words of the input problems. We found that the model usually attends to a single word to make its prediction, irrespective of the context. Table \ref{tab:attn_wts} shows some representative examples. In the first example, the model assigns an attention weight of 1 to the representation of the word \textit{`every'} and predicts the correct equation. However, when we make a subtle change to this problem such that the corresponding equation changes, the model keeps on attending over the word \textit{`every'} and predicts the same equation, which is now incorrect. Similar observations can be made for the other two examples. Table \ref{tab:attn_wts_sup} in the Appendix has more such examples. These examples represent only a few types of spurious correlations that we could find but there could be other types of correlations that might have been missed.


Note that, we do not claim that every model trained on these datasets relies on the occurrence of specific words in the input problem for prediction the way our constrained model does. We are only asserting that it is possible to achieve a good score on these datasets even with such a brittle model, which clearly makes these datasets unreliable to robustly measure model performance.

\section{\dataname{}}


The efficacy of existing models on benchmark datasets has led to a shift in the focus of researchers towards more difficult MWPs. We claim that this efficacy on benchmarks is misleading and SOTA MWP solvers are unable to solve even elementary level one-unknown MWPs. To this end, we create a challenge set named \dataname{} containing simple one-unknown arithmetic word problems of grade level up to 4. The examples in \dataname{} test a model across different aspects of solving word problems. For instance, a model needs to be sensitive to questions and possess certain reasoning abilities to correctly solve the examples in our challenge set. \dataname{} is similar to existing datasets of the same level in terms of scope and difficulty for humans, but is less susceptible to being solved by models relying on superficial patterns.

Our work differs from adversarial data collection methods such as Adversarial NLI \cite{anli} in that these methods create examples depending on the failure of a particular model while we create examples without referring to any specific model. Inspired by the notion of \emph{Normative evaluation} \cite{linzen-2020-accelerate}, our goal is to create a dataset of simple problems that any system designed to solve MWPs should be expected to solve. We create new problems by applying certain variations to existing problems, similar to the work of \citet{checklist}. However, unlike their work, our variations do not check for linguistic capabilities. Rather, the choice of our variations is motivated by the experiments in Section~\ref{flaws} as well as certain simple capabilities that any MWP solver must possess.



\begin{table}[b]
	\small{\centering
		\begin{tabular}{p{6em}P{6em}P{6em}}
			\toprule
			\textbf{Group} & \textbf{Examples in ASDiv-A}& \textbf{Selected Seed Examples}\\
			\midrule
			Addition & 278 & 28 \\
			Subtraction & 362 & 33 \\
			Multiplication & 188 & 19 \\
			Division & 176 & 20 \\
			\midrule
			Total & 1004 & 100 \\
			\bottomrule
		\end{tabular}
		\caption{\label{tab:seed_dist}Distribution of selected seed examples across types.}
	}
\end{table}
\definecolor{carminepink}{rgb}{0.88, 0.4, 0.56}
\setlength{\extrarowheight}{2pt}
\begin{table*}[t]
	\small{\centering
		\begin{tabular}{>{\raggedright}m{5em}>{\raggedright}m{8em}m{32em}}
			\toprule
			\textsc{Category} & \textsc{Variation} & \textsc{Examples}\\
			\midrule
			\multirow{7}{=}{\footnotesize{Question Sensitivity}} &
			\scriptsize{Same Object, Different Structure} &
			\scriptsize{\textbf{\textcolor{brown}{Original:}} Allan brought two balloons and Jake brought four balloons to the park. How many \textcolor{carminepink}{balloons} did Allan and Jake have in the park? \newline \textbf{\textcolor{violet}{Variation:}} Allan brought two balloons and Jake brought four balloons to the park. How many \textcolor{green!45!blue}{more} \textcolor{carminepink}{balloons} did \textcolor{green!45!blue}{Jake have than Allan} in the park?} \\
			\cline{2-3}
			& \scriptsize{Different Object, Same Structure} &
			\scriptsize{\textbf{\textcolor{brown}{Original:}} 
				In a school, there are 542 girls and 387 boys. 290 more boys joined the school. How many \textcolor{carminepink}{pupils} are in the school?
				\newline \textbf{\textcolor{violet}{Variation:}} 
				In a school, there are 542 girls and 387 boys. 290 more boys joined the school. How many \textcolor{carminepink}{boys} are in the school?
			} \\
			\cline{2-3}
			& \scriptsize{Different Object, Different Structure} &
			\scriptsize{\textbf{\textcolor{brown}{Original:}}  
				He then went to see the oranges being harvested. He found out that they harvest 83 sacks per day and that each sack contains 12 oranges. How many \textcolor{carminepink}{sacks of oranges} will they have after 6 days of harvest?
				\newline \textbf{\textcolor{violet}{Variation:}} 
				He then went to see the oranges being harvested. He found out that they harvest 83 sacks per day and that each sack contains 12 oranges. How many \textcolor{carminepink}{oranges} \textcolor{green!45!blue}{do they harvest per day}?
			}\\
			
			\midrule
			
			\multirow{7}{=}{\footnotesize{Reasoning Ability}} &
			\scriptsize{Add relevant information} &
			\scriptsize{\textbf{\textcolor{brown}{Original:}} Every day, Ryan spends 4 hours on learning English and 3 hours on learning Chinese. How many hours does he spend on learning English and Chinese in all? \newline \textbf{\textcolor{violet}{Variation:}} Every day, Ryan spends 4 hours on learning English and 3 hours on learning Chinese. \textcolor{green!45!blue}{If he learns for 3 days,} how many hours does he spend on learning English and Chinese in all?}\\
			\cline{2-3}
			& \scriptsize{Change Information} &
			\scriptsize{\textbf{\textcolor{brown}{Original:}}  \textcolor{carminepink}{Jack} had 142 pencils. Jack gave 31 pencils to Dorothy. How many pencils does \textcolor{carminepink}{Jack} have now? \newline \textbf{\textcolor{violet}{Variation:}} \textcolor{green!45!blue}{Dorothy} had 142 pencils. Jack gave 31 pencils to Dorothy. How many pencils does \textcolor{green!45!blue}{Dorothy} have now?} \\
			\cline{2-3}
			& \scriptsize{Invert Operation} &
			\scriptsize{\textbf{\textcolor{brown}{Original:}} He also made some juice from fresh oranges. If he used 2 oranges per glass of juice and he made \textcolor{green!45!blue}{6 glasses of juice}, \textcolor{carminepink}{how many oranges} did he use? \newline \textbf{\textcolor{violet}{Variation:}} He also made some juice from fresh oranges. If he used 2 oranges per glass of juice and he used up \textcolor{carminepink}{12 oranges}, \textcolor{green!45!blue}{how many glasses of juice} did he make?}\\
			
			\midrule
			
			\multirow{6}{=}{\footnotesize{Structural Invariance} } &
			\scriptsize{Change order of objects} &
			\scriptsize{\textbf{\textcolor{brown}{Original:}} 
				John has \textcolor{carminepink}{8 marbles} and \textcolor{green!45!blue}{3 stones}. How many more marbles than stones does he have?
				\newline \textbf{\textcolor{violet}{Variation:}} 
				John has \textcolor{green!45!blue}{3 stones} and \textcolor{carminepink}{8 marbles}. How many more marbles than stones does he have?
			} \\
			\cline{2-3}
			& \scriptsize{Change order of phrases} &
			\scriptsize{\textbf{\textcolor{brown}{Original:}} 
				\textcolor{carminepink}{Matthew had 27 crackers}. If \textcolor{green!45!blue}{Matthew gave equal numbers of crackers to his 9 friends}, how many crackers did each person eat?
				\newline \textbf{\textcolor{violet}{Variation:}} \textcolor{green!45!blue}{Matthew gave equal numbers of crackers to his 9 friends}. If \textcolor{carminepink}{Matthew had a total of 27 crackers initially}, how many crackers did each person eat?} \\
			\cline{2-3}
			& \scriptsize{Add irrelevant information} &
			\scriptsize{\textbf{\textcolor{brown}{Original:}} Jack had 142 pencils. Jack gave 31 pencils to Dorothy. How many pencils does Jack have now? \newline \textbf{\textcolor{violet}{Variation:}} Jack had 142 pencils. \textcolor{green!45!blue}{Dorothy had 50 pencils.} Jack gave 31 pencils to Dorothy. How many pencils does Jack have now?}\\
			\bottomrule
		\end{tabular}
		\caption{\label{tab:var_types} Types of Variations with examples. `\textbf{\textcolor{brown}{Original:}}' denotes the base example from which the variation is created, `\textbf{\textcolor{violet}{Variation:}}' denotes a manually created variation.}
	}
\end{table*}
\setlength{\extrarowheight}{0pt}
\subsection{Creating \dataname{}}

We create \dataname{} by applying certain types of variations to a set of seed examples sampled from the ASDiv-A dataset. We select the seed examples from the recently proposed ASDiv-A dataset since it appears to be of higher quality and harder than the MAWPS dataset: We perform a simple experiment to test the coverage of each dataset by training a model on one dataset and testing it on the other one. For instance, when we train a Graph2Tree model on ASDiv-A, it achieves 82\% accuracy on MAWPS. However, when trained on MAWPS and tested on ASDiv-A, the model achieved only 73\% accuracy. Also recall Table~\ref{tab:std_scores} where most models performed better on MAWPS. Moreover, ASDiv has problems annotated according to types and grade levels which are useful for us. 

To select a subset of seed examples that sufficiently represent different types of problems in the ASDiv-A dataset, we first divide the examples into groups according to their annotated types. We discard types such as \textit{`TVQ-Change'}, \textit{`TVQ-Initial'}, \textit{`Ceil-Division'} and \textit{`Floor-Division'} that have less than 20 examples each. We also do not consider the \textit{`Difference'} type since it requires the use of an additional modulus operator.  For ease of creation, we discard the few examples that are more than 40 words long. To control the complexity of resulting variations, we only consider those problems as seed examples that can be solved by an expression with a single operator. Then, within each group, we cluster examples using  K-Means over RoBERTa sentence embeddings of each example. From each cluster, the example closest to the cluster centroid is selected as a seed example. We selected a total of 100 seed examples in this manner. The distribution of seed examples according to different types of problems can be seen in Table \ref{tab:seed_dist}.

\subsubsection{Variations}\label{var_types}

 The variations that we make to each seed example can be broadly classified into three categories based on desirable properties of an ideal model: \textit{Question Sensitivity}, \textit{Reasoning Ability} and \textit{Structural Invariance}. Examples of each type of variation are provided in Table \ref{tab:var_types}.\\

	\noindent\textbf{1. Question Sensitivity.}		
	\noindent Variations in this category check if the model's answer depends on the question. In these variations, we change the question in the seed example while keeping the body same. The possible variations are as follows:
	
	\noindent\emph{(a) Same Object, Different Structure:} The principal object (i.e. object whose quantity is unknown) in the question is kept the same while the structure of the question is changed.
	
	\noindent\emph{(b) Different Object, Same Structure:} The principal object in the question is changed while the structure of question remains fixed.
	
	\noindent\emph{(c) Different Object, Different Structure:} Both, the principal object in the question and the structure of the question, are changed.\\
	
%
%
%
%
%
%
	\noindent\textbf{2. Reasoning Ability.}	
	\noindent Variations here check whether a model has the ability to correctly determine a change in reasoning arising from subtle changes in the problem text. The different possible variations are as follows:
	
	\noindent\emph{(a) Add relevant information:} Extra relevant information is added to the
	example that affects the output equation.
	
	\noindent\emph{(b) Change information:} The information provided in the example is changed.
	
	\noindent\emph{(c) Invert operation:} The previously unknown quantity is now provided as information and the question instead asks about a previously known quantity which is now unknown.\\
	
%
%
%
%
%
%

	\noindent\textbf{3. Structural Invariance.}
	\noindent Variations in this category check whether a model remains invariant to superficial structural changes that do not alter the answer or the reasoning required to solve the example. The different possible variations are as follows:
	
	\noindent\emph{(a) Add irrelevant information:} Extra irrelevant information is added to the problem text that is not required to solve the example.
	
	\noindent\emph{(b) Change order of objects:} The order of objects appearing in the example
	is changed.
	
	\noindent\emph{(c) Change order of phrases:} The order of number-containing phrases appearing in the example is changed.
	
%
%
%
%
%
%
	

\begin{table}[t]
	\scriptsize{\centering
		\begin{tabular}{p{4em}P{5em}P{5em}P{4.5em}P{3em}}
			\toprule
			\textbf{Dataset} & \textbf{\# Problems}& \textbf{\# Equation Templates} & \textbf{\# Avg Ops} & \textbf{CLD}\\
			\midrule
			MAWPS & 2373 & 39 & 1.78 & 0.26 \\
			ASDiv-A & 1218 & 19 & 1.23 & 0.50\\
			\dataname{} & 1000 & 26 & 1.24 & 0.22\\
			\bottomrule
		\end{tabular}
		\caption{\label{tab:data_stats}Statistics of our dataset compared with MAWPS and ASDiv-A.}
	}
\end{table}

\subsubsection{Protocol for creating variations}

Since creating variations requires a high level of familiarity with the task, the construction of \dataname{} is done in-house by the authors and colleagues, hereafter called the \textit{workers}. The 100 seed examples (as shown in Table \ref{tab:seed_dist}) are distributed among the \textit{workers}.



For each seed example, the \textit{worker} needs to create new variations by applying the variation types discussed in Section \ref{var_types}. Importantly, a combination of different variations over the seed example can also be done. For each new example created, the worker needs to annotate it with the equation as well as the type of variation(s) used to create it. More details about the creation protocol can be found in Appendix \ref{asec:protocol}.

We created a total of 1098 examples. However, since ASDiv-A does not have examples with equations of more than two operators, we discarded 98 examples from our set which had equations consisting of more than two operators. This is to ensure that our challenge set does not have any unfairly difficult examples. The final set of 1000 examples was provided to an external volunteer unfamiliar with the task to check the grammatical and logical correctness of each example. 

\subsection{Dataset Properties}\label{data_props}

Our challenge set \dataname{} consists of one-unknown arithmetic word problems which can be solved by expressions requiring no more than two operators.
Table \ref{tab:data_stats} shows some statistics of our dataset and of ASDiv-A and MAWPS. The Equation Template for each example is obtained by converting the corresponding equation into prefix form and masking out all numbers with a meta symbol. Observe that the number of distinct Equation Templates and the Average Number of Operators are similar for \dataname{} and ASDiv-A and are considerably smaller than for MAWPS. This indicates that \dataname{} does not contain unfairly difficult MWPs in terms of the arithmetic expression expected to be produced by a model.

Previous works, including those introducing MAWPS and ASDiv, have tried to capture the notion of \textit{diversity} in MWP datasets. \citet{miao-etal-2020-diverse} introduced a metric called Corpus Lexicon Diversity (CLD) to measure lexical diversity. Their contention was that higher lexical diversity is correlated with the quality of a dataset. As can be seen from Table \ref{tab:data_stats}, \dataname{} has a much lesser CLD than ASDiv-A.
\dataname{} is also less diverse in terms of problem types compared to ASDiv-a. Despite this we will show in the next section that \dataname{} is in fact more challenging than ASDiv-A for current models. Thus, we believe that lexical diversity is not a reliable way to measure the quality of MWP datasets. Rather it could depend on other factors such as the diversity in MWP structure which preclude models exploiting shallow heuristics.

\subsection{Experiments on \dataname{}}

We train the three considered models on a combination of MAWPS and ASDiv-A and test them on \dataname{}. The scores of all three models with and without RoBERTa embeddings for various subsets of \dataname{} can be seen in Table \ref{tab:chall_scores}.

\begin{table}[t]
	\footnotesize{\centering
		\begin{tabular}{p{4em}P{1.7em}P{1.7em}P{1.7em}P{1.7em}P{1.7em}P{1.7em}}
			\toprule
			&\multicolumn{2}{c}{\textbf{Seq2Seq}} &\multicolumn{2}{c}{\textbf{GTS}}
			&\multicolumn{2}{c}{\textbf{Graph2Tree}} \\ 
			\cmidrule(lr){2-3}\cmidrule(lr){4-5}\cmidrule(lr){6-7}
			& \textit{S} & \textit{R} & \textit{S} & \textit{R} & \textit{S} & \textit{R} \\
			\midrule
			Full Set & 24.2 & 40.3 & 30.8 & 41.0 & 36.5 & \textbf{43.8} \\
			\midrule
			One-Op & 25.4 & 42.6 & 31.7 & 44.6 & 42.9 & \textbf{51.9} \\
			Two-Op & 20.3 & \textbf{33.1} & 27.9 & 29.7 & 16.1 & 17.8 \\
			\midrule
			ADD & 28.5 & \textbf{41.9} & 35.8 & 36.3 & 24.9 & 36.8 \\
			SUB & 22.3 & 35.1 & 26.7 & 36.9 & 41.3 & \textbf{41.3} \\
			MUL & 17.9 & \textbf{38.7} & 29.2 & \textbf{38.7} & 27.4 & 35.8 \\
			DIV & 29.3 & 56.3 & 39.5 & 61.1 & 40.7 & \textbf{65.3} \\
			\bottomrule
		\end{tabular}
		\caption{\label{tab:chall_scores}Results of models on the \dataname{} challenge set. \textit{S} indicates that the model is trained from scratch. \textit{R}  indicates that the model was trained with RoBERTa embeddings. The first row shows the results for the full dataset. The next two rows show the results for subsets of \dataname{} composed of examples that have equations with one operator and two operators respectively. The last four rows show the results for subsets of \dataname{} composed of examples of type Addition, Subtraction, Multiplication and Division respectively.}
	}
\end{table}

\begin{table}[b]
	\small{\centering
		\begin{tabular}{p{4em}P{8em}P{8em}}
			\toprule
			\textbf{Model} & \textbf{\dataname{} w/o ques} & \textbf{ASDiv-A w/o ques}\\
			\midrule
			Seq2Seq & 29.2 & 58.7 \\
			GTS & 28.6 & 60.7 \\
			Graph2Tree & \textbf{30.8} & \textbf{64.4} \\
			\bottomrule
		\end{tabular}
		\caption{\label{tab:prem-only_chall}Accuracies ($\uparrow$) of models on \dataname{} without questions. The 5-fold CV accuracy scores for ASDiv-A without questions are restated for easier comparison.}
	}
\end{table}

The best performing Graph2Tree model is only able to achieve an accuracy of 43.8\% on \dataname{}. This indicates that the problems in \dataname{} are indeed more challenging for the models than the problems in ASDiv-A and MAWPS despite being of the same scope and type and less diverse. Table \ref{tab:failure_examples} in the Appendix lists some simple examples from SVAMP on which the best performing model fails. These results lend further support to our claim that existing models cannot robustly solve elementary level word problems.


Next, we remove the questions from the examples in \dataname{} and evaluate them using the three models with RoBERTa embeddings trained on combined MAWPS and ASDiv-A. The scores can be seen in Table \ref{tab:prem-only_chall}. The accuracy drops by half when compared to ASDiv-A and more than half compared to MAWPS suggesting that the problems in \dataname{} are more sensitive to the information present in the question. We also evaluate the performance of the constrained model on \dataname{} when trained on MAWPS and ASDiv-A. The best model achieves only 18.3\% accuracy (see Table \ref{tab:simple_chall}) which is marginally better than the majority template baseline. This shows that the problems in \dataname{} are less vulnerable to being solved by models using simple patterns and that a model needs contextual information in order to solve them.

We also explored using \dataname{} for training by combining it with ASDiv-A and MAWPS. We performed 5-fold cross-validation over \dataname{} where the model was trained on a combination of the three datasets and tested on unseen examples from \dataname{}. To create the folds, we first divide the seed examples into five sets, with each \emph{type} of example distributed nearly equally among the sets. A fold is obtained by combining all the examples in SVAMP that were created using the seed examples in a set. In this way, we get five different folds from the five sets.  We found that the best model achieved about 65\% accuracy. This indicates that even with additional training data existing models are still not close to the performance that was estimated based on prior benchmark datasets.

\begin{table}[t]
	\small{\centering
		\begin{tabular}{p{11em}P{6em}}
			\toprule
			\textbf{Model} & \textbf{\dataname{}}\\
			\midrule
			FFN + LSTM Decoder (S) & 17.5 \\
			FFN + LSTM Decoder (R) & \textbf{18.3} \\
			\midrule
			Majority Template Baseline & 11.7 \\
			\bottomrule
		\end{tabular}
		\caption{\label{tab:simple_chall} Accuracies ($\uparrow$) of the constrained model on \dataname{}. (R) denotes that the model is provided with non-contextual RoBERTa pretrained embeddings while (S) denotes that the model is trained from scratch.}
	}
\end{table}


\begin{table}[b]
	\small{\centering
		\begin{tabular}{p{8.5em}P{5em}P{6em}}
			\toprule
			\textbf{Removed Category} & \textbf{\# Removed Examples} & \textbf{Change in Accuracy $(\Delta)$}\\
			\midrule
			Question Sensitivity & 462 & +13.7 \\
			Reasoning Ability & 649 & -3.3 \\
			Structural Invariance & 467 & +4.5\\
			\bottomrule
		\end{tabular}
		\caption{\label{tab:cat_wise}Change in accuracies when categories are removed. The Change in Accuracy $\Delta = Acc(Full - Cat) - Acc(Full)$, where $Acc(Full)$ is the accuracy on the full set and $Acc(Full - Cat)$ is the accuracy on the set of examples left after removing all examples which were created using Category $Cat$ either by itself, or in use with other categories.}
	}
\end{table}

\begin{table}[t]
	\small{\centering
		\begin{tabular}{p{9.5em}P{5em}P{6em}}
			\toprule
			\textbf{Removed Variation} & \textbf{\# Removed Examples} & \textbf{Change in Accuracy $(\Delta)$}\\
			\midrule
			Same Obj, Diff Struct & 325 & +7.3 \\
			Diff Obj, Same Struct & 69 & +1.5 \\
			Diff Obj, Diff Struct & 74 & +1.3 \\
			\midrule
			Add Rel Info & 264 & +5.5 \\
			Change Info & 149 & +3.2 \\
			Invert Operation & 255 & -10.2 \\
			\midrule
			Change order of Obj & 107 & +2.3 \\
			Change order of Phrases & 152 & -3.3 \\
			Add Irrel Info & 281 & +6.9 \\
			\bottomrule
		\end{tabular}
		\caption{\label{tab:var_wise}Change in accuracies when variations are removed. The Change in Accuracy $\Delta = Acc(Full - Var) - Acc(Full)$, where $Acc(Full)$ is the accuracy on the full set and $Acc(Full - Var)$ is the accuracy on the set of examples left after removing all examples which were created using Variation $Var$ either by itself, or in use with other variations.}
	}
\end{table}

To check the influence of different categories of variations in \dataname{}, for each category, we measure the difference between the accuracy of the best model on the full dataset and its accuracy on a subset containing no example created from that category of variations. The results are shown in Table \ref{tab:cat_wise}. Both the \textit{Question Sensitivity} and \textit{Structural Invariance} categories of variations show an increase in accuracy when their examples are removed, thereby indicating that they make \dataname{} more challenging. The decrease in accuracy for the \textit{Reasoning Ability} category can be attributed in large part to the \textit{Invert Operation} variation. This is not surprising because most of the examples created from \textit{Invert Operation} are almost indistinguishable from examples in ASDiv-A, which the model has seen during training. The scores for each individual variation are provided in Table \ref{tab:var_wise}.

We also check the break-up of performance of the best performing Graph2Tree model according to the number of numbers present in the text of the input problem. We trained the model on both ASDiv-A and MAWPS and tested on \dataname{} and compare those results against the 5-fold cross-validation setting of ASDiv-A. The scores are provided in Table \ref{tab:num_nums}. While the model can solve many problems consisting of only two numbers in the input text (even in our challenge set), it performs very badly on problems having more than two numbers. This shows that current methods are incapable of properly associating numbers to their context. Also, the gap between the performance on ASDiv-A and \dataname{} is high, indicating that the examples in \dataname{} are more difficult for these models to solve than the examples in ASDiv-A even when considering the structurally same type of word problems.

\begin{table}[t]
	\small{\centering
		\begin{tabular}{p{4em}P{4em}P{4em}P{4em}}
			\toprule
			\textbf{Dataset} & \textbf{2 nums} & \textbf{3 nums} & \textbf{4 nums}\\
			\midrule
			ASDiv-A & 93.3 & 59.0 & 47.5 \\
			\dataname{} & 78.3 & 25.4 & 25.4 \\
			\bottomrule
		\end{tabular}
		\caption{\label{tab:num_nums}Accuracy break-up according to the number of numbers in the input problem. \textbf{2 nums} refers to the subset of problems which have only 2 numbers in the problem text. Similarly, \textbf{3 nums} and \textbf{4 nums} are subsets that contain 3 and 4 different numbers in the problem text respectively.}
	}
\end{table}

\section{Final Remarks}

Going back to the original question, are existing NLP models able to solve elementary math word problems? This paper gives a negative answer. We have empirically shown that the benchmark English MWP datasets suffer from artifacts making them unreliable to gauge the performance of MWP solvers: we demonstrated that the majority of problems in the existing datasets can be solved by simple heuristics even without word-order information or the question text. 

The performance of the existing models in our proposed challenge dataset also highlights their limitations in solving simple elementary level word problems.  We hope that our challenge set \dataname{}, containing elementary level MWPs, will enable more robust evaluation of methods. We believe that methods proposed in the future that make genuine advances in solving the task rather than relying on simple heuristics will perform well on \dataname{} despite being trained on other datasets such as ASDiv-A and MAWPS.

In recent years, the focus of the community has shifted towards solving more difficult MWPs such as non-linear equations and word problems with multiple unknown variables. We demonstrated that the capability of existing models to solve simple one-unknown arithmetic word problems is overestimated. We believe that developing more robust methods for solving elementary MWPs remains a significant open problem.

\section*{Acknowledgements}

We thank the anonymous reviewers for their constructive comments.  We would also like to thank our colleagues at Microsoft Research for providing valuable feedback. We are grateful to Monojit Choudhury for discussions about creating the dataset. We thank Kabir Ahuja for carrying out preliminary experiments that led to this work. We also thank Vageesh Chandramouli and Nalin Patel for their help in dataset construction.

\bibliography{citations}
\bibliographystyle{acl_natbib}

\clearpage
\newpage
\appendix

\section{Experiments with Transformer}

We additionally ran all our experiments with the Transformer \cite{transformer} model. The 5-fold cross-validation accuracies of the Transformer on MAWPS and ASDiv-A are provided in Table \ref{tab:trans_std_scores}. The scores on Question-removed datasets are provided in Table \ref{tab:trans_prem-only_scores} and on \dataname{} challenge set is provided in Table \ref{tab:trans_chall_scores}.

\begin{table}[b]
	\small{\centering
		\begin{tabular}{p{10em}P{3.5em}P{4em}}
			\toprule
			\textbf{Model} & \textbf{MAWPS}& \textbf{ASDiv-A}\\
			\midrule
			Transformer (S)  & 77.9 & 52.1 \\
			Transformer (R) & 87.1 & 77.7 \\
			\bottomrule
		\end{tabular}
		\caption{\label{tab:trans_std_scores}5-fold cross-validation accuracies ($\uparrow$) of Transformer model on datasets. (R) means that the model is provided with RoBERTa pretrained embeddings while (S) means that the model is trained from scratch.}
	}
\end{table}

\begin{table}[h]
	\small{\centering
		\begin{tabular}{p{6em}P{4em}P{4em}P{4em}}
			\toprule
			\textbf{Model} & \textbf{MAWPS} & \textbf{ASDiv-A} & \textbf{SVAMP} \\
			\midrule
			Transformer & 79.4 & 64.4 & 25.3 \\
			\bottomrule
		\end{tabular}
		\caption{\label{tab:trans_prem-only_scores}5-fold cross-validation accuracies ($\uparrow$) of Transformer model on Question-removed datasets.}
	}
\end{table}

\begin{table}[h]
	\footnotesize{\centering
		\begin{tabular}{p{6em}P{4em}P{4em}}
			\toprule
			&\multicolumn{2}{c}{\textbf{Transformer}} \\ 
			\cmidrule(lr){2-3}
			& \textit{S} & \textit{R} \\
			\midrule
			Full Set & 18.4 & 38.9 \\
			\midrule
			One-Op & 18.6 & 40.5 \\
			Two-Op & 17.8 & 33.9 \\
			\midrule
			ADD & 22.3 & 36.3 \\
			SUB & 17.1 & 37.5 \\
			MUL & 17.9 & 28.3 \\
			DIV & 18.6 & 53.3 \\
			\bottomrule
		\end{tabular}
		\caption{\label{tab:trans_chall_scores}Results of Transformer model on the \dataname{} challenge set. \textit{S} indicates that the model is trained from scratch. \textit{R}  indicates that the model was trained with RoBERTa embeddings. The first row shows the results for the full dataset. The next two rows show the results for subsets of \dataname{} composed of examples that have equations with one operator and two operators respectively. The last four rows show the results for subsets of \dataname{} composed of examples of type Addition, Subtraction, Multiplication and Division respectively.}
	}
\end{table}

\section{Implementation Details}\label{implementation}

\begin{table*}[h]
	\scriptsize{\centering
		\begin{tabular}{p{7.5em}P{5em}P{5em}P{5em}P{5em}P{5em}P{5em}P{5em}P{5em}}
			\toprule
			&\multicolumn{2}{c}{\textbf{Seq2Seq}} &\multicolumn{2}{c}{\textbf{GTS}}
			&\multicolumn{2}{c}{\textbf{Graph2Tree}}
			&\multicolumn{2}{c}{\textbf{Constrained}} \\ 
			\cmidrule(lr){2-3}\cmidrule(lr){4-5}\cmidrule(lr){6-7}\cmidrule(lr){8-9}
			\textbf{Hyperparameters} & Scratch& RoBERTa & Scratch& RoBERTa & Scratch& RoBERTa & Scratch& RoBERTa\\
			\midrule
			Embedding Size & [128, \textbf{256}] & [768] & [\textbf{128}, 256] & [768] & [\textbf{128}, 256] & [768] & [\textbf{128}, 256] & [768]\\
			
			Hidden Size & [\textbf{256}, 384] & [\textbf{256}, 384] & [384, \textbf{512}] & [384, \textbf{512}] & [256, \textbf{384}] & [256, \textbf{384}] & [256, \textbf{384}] & [\textbf{256}, 384]\\
			
			Number of Layers & [1, \textbf{2}] & [1, \textbf{2}] & [1, \textbf{2}] & [1, \textbf{2}] & [1, \textbf{2}] & [1, \textbf{2}] & [\textbf{1}, 2] & [\textbf{1}, 2]\\
			
			Learning Rate & [5e-4, 8e-4, \textbf{1e-3}] & [1e-4, \textbf{2e-4}, 5e-4] & [8e-4, 1e-3, \textbf{2e-3}] & [5e-4, 8e-4, \textbf{1e-3}] & [\textbf{8e-4}, 1e-3, 2e-3] & [5e-4, \textbf{8e-4}, 1e-3] & [\textbf{1e-3}, 2e-3] & [1e-3, \textbf{2e-3}]\\
			
			Embedding LR & [5e-4, \textbf{8e-4}, 1e-3] & [5e-6, \textbf{8e-6}, 1e-5] & [8e-4, 1e-3, \textbf{2e-3}] & [5e-6, \textbf{8e-6}, 1e-5] & [\textbf{8e-4}, 1e-3, 2e-3] & [5e-6, 8e-6, \textbf{1e-5}] & [\textbf{1e-3}, 2e-3] & [1e-3, \textbf{2e-3}]\\
			
			Batch Size & [\textbf{8}, 16] & [\textbf{4}, 8] & [8, \textbf{16}] & [\textbf{4}, 8] & [\textbf{8}, 16] & [4, \textbf{8}] & [8, \textbf{16}] & [\textbf{4}, 8] \\
			
			Dropout & [0.1] & [0.1] & [0.5] & [0.5] & [0.5] & [0.5] & [0.1] & [0.1]\\
			
			\midrule
			
			\# Parameters & 8.5M & 130M & 15M & 140M & 16M & 143M & 5M  & 130M \\
			
			Epochs & 60 & 50 & 60 & 50 & 60 & 50 & 60 & 50 \\
			
			Avg Time/Epoch & 10 & 40 & 60 & 120 & 60 & 120 & 10 & 15 \\
			
			\bottomrule
		\end{tabular}
		\caption{\label{tab:hyperparams}Different hyperparameters and the values considered for each of them in the models. The best hyperparameters for each model for 5-fold cross-validation on ASDiv-A are highlighted in bold. Average Time/Epoch is measured in seconds.}
	}
\end{table*}

\begin{table}[h]
	\scriptsize{\centering
		\begin{tabular}{p{11.5em}P{7em}P{7em}}
			\toprule
			&\multicolumn{2}{c}{\textbf{Transformer}} \\ 
			\cmidrule(lr){2-3}
			\textbf{Hyperparameters} & Scratch & RoBERTa \\
			\midrule
			I/P and O/P Embedding Size & [\textbf{128}, 256] & [768] \\
			FFN Size & [\textbf{256}, 384] & [\textbf{256}, 384] \\
			heads & [2, \textbf{4}] & [2, \textbf{4}] \\
			Number of Encoder Layers & [\textbf{1}, 2] & [\textbf{1}, 2] \\
			Number of Decoder Layers & [\textbf{1}, 2] & [\textbf{1}, 2] \\
			Learning Rate & [5e-5, 8e-5, \textbf{1e-4}] & [5e-5, 8e-5, \textbf{1e-4}] \\
			Embedding LR & [5e-5, 8e-5, \textbf{1e-4}] & [1e-5, \textbf{5e-6}] \\
			Batch Size & [\textbf{4}, 8] & [\textbf{4}, 8] \\
			Dropout & [0.1] & [0.1] \\
			\midrule
			\# Parameters & 0.67M & 132M \\
			Epochs & 100 & 100 \\
			Avg Time/Epoch & 10 & 30 \\
			
			\bottomrule
		\end{tabular}
		\caption{\label{tab:trans_hyperparams}Different hyperparameters and the values considered for each of them in the Transformer model. The best hyperparameters for 5-fold cross-validation on ASDiv-A are highlighted in bold. Average Time/Epoch is measured in seconds.}
	}
\end{table}

We use 8 NVIDIA Tesla P100 GPUs each with 16 GB memory to run our experiments. The hyperparameters used for each model are shown in Table \ref{tab:hyperparams}. The hyperparameters used in for the Transformer model are provided in Table \ref{tab:trans_hyperparams}. The best hyperparameters are highlighted in bold. Following the setting of \citet{zhang-etal-2020-graph}, the arithmetic word problems from MAWPS are divided into five folds, each of equal test size. For ASDiv-A, we consider the 5-fold split [238, 238, 238, 238, 266] provided by the authors \cite{miao-etal-2020-diverse}. 

\section{Creation Protocol}\label{asec:protocol}

\begin{table}[t]
	\small{\centering
		\begin{tabular}{P{8em}P{10em}}
			\toprule
			
			\textbf{Tag} & \textbf{Description}\\
			\midrule
			\big[NUM\textbf{\textit{x}}\big] & Number\\
			\big[NAME\textbf{\textit{x}}\big] & Names of Persons\\
			\big[OBJs\textbf{\textit{x}}\big] & Singular Object\\
			\big[OBJp\textbf{\textit{x}}\big] & Plural Object\\
			\big[MOD\textbf{\textit{x}}\big] & Modifier\\
			\bottomrule
		\end{tabular}
		
		\caption{\label{tab:tags} List of tags used in annotated templates. \textbf{\textit{x}} denotes the index of the tag.}
	}
\end{table}

We create variations in template form. Generating more data by scaling up from these templates or by performing automatic operations on these templates is left for future work. The template form of an example is created by replacing certain words with their respective tags. Table \ref{tab:tags} lists the various tags used in the templates.

The $\big[NUM\big]$ tag is used to replace all the numbers and the $\big[NAME\big]$ tag is used to replace all the Names of Persons in the example. The $\big[OBJs\big]$ and $\big[OBJp\big]$ tags are used for replacing the objects in the example. The $\big[OBJs\big]$ and $\big[OBJp\big]$ tags with the same index represent the same object in singular and plural form respectively. The intention when using the $\big[OBJs\big]$ or the $\big[OBJp\big]$ tag is that it can be used as a placeholder for other similar words, which when entered in that place, make sense as per the context. These tags must not be used for collectives; rather they should be used for the things that the collective represents. Some example uses of $\big[OBJs\big]$ and $\big[OBJp\big]$ tags are provided in Table \ref{tab:obj_examples}. Lastly, the $\big[MOD\big]$ tag must be used to replace any modifier preceding the $\big[OBJs\big]$/$\big[OBJp\big]$ tag.

A preprocessing script is executed over the \textbf{\textit{Seed Examples}} to automatically generate template suggestions for the \textit{workers}. The script uses Named Entity Recognition and Regular Expression matching to automatically mask the names of persons and the numbers found in the \textbf{\textit{Seed Examples}}. The outputs from the script are called the \textbf{\textit{Script Examples}}. An illustration is provided in Table \ref{tab:template_examples}.

Each \textit{worker} is provided with the \textbf{\textit{Seed Examples}} along with their respective \textbf{\textit{Script Examples}} that have been alloted to them. The \textit{worker's} task is to edit the \textbf{\textit{Script Example}} by correcting any mistake made by the preprocessing script and adding any new tags such as the $\big[OBJs\big]$ and the $\big[OBJp\big]$ tags in order to create the \textbf{\textit{Base Example}}. If a \textit{worker} introduces a new tag, they need to mark it against its example-specific value. If the tag is used to mask objects, the \textit{worker} needs to mark both the singular and plural form of the object in a comma-seperated manner. Additionally, for each unique index of $\big[OBJs\big]$/$\big[OBJp\big]$ tag in the example, the \textit{worker} must enter atleast one alternate value that can be used in that place. Similarly, the \textit{worker} must enter atleast two modifier words that can be used to precede the principal $\big[OBJs\big]$/$\big[OBJp\big]$ tags in the example. These alternate values are used to gather a lexicon which can be utilised to scale-up the data at a later stage. An illustration of this process is provided in Table \ref{tab:og_examples}.

In order to create the variations, the \textit{worker} needs to check the different types of variations in Table \ref{tab:var_types} to see if they can be applied to the \textbf{\textit{Base Example}}. If applicable, the \textit{worker} needs to create the \textbf{\textit{Variation Example}} while also making a note of the type of variation. If a particular example is the result of performing multiple types of variations, all types of variations should be listed according to their order of application from latest to earliest in a comma-seperated manner. For any variation, if a \textit{worker} introduces a new tag, they need to mark it against its example-specific value as mentioned before. The index of any new tag introduced needs to be one more than the highest index already in use for that tag in the \textbf{\textit{Base Example}} or its previously created variations.

To make the annotation more efficient and streamlined, we provide the following steps to be followed in order:
\begin{enumerate}
	\item Apply the \textit{Question Sensitivity} variations on the \textbf{\textit{Base Example}}.
	\item Apply the \textit{Invert Operation} variation on the \textbf{\textit{Base Example}} and on all the variations obtained so far.
	\item Apply the \textit{Add relevant information} variation on the \textbf{\textit{Base Example}}. Then considering these variations as \textbf{\textit{Base Examples}}, apply the \textit{Question Sensitivity} variations.
	\item Apply the \textit{Add irrelevant information} variation on the \textbf{\textit{Base Example}} and on all the variations obtained so far.
	\item Apply the \textit{Change information} variation on the \textbf{\textit{Base Example}} and on all the variations obtained so far.
	\item Apply the \textit{Change order of Objects} and \textit{Change order of Events or Phrases} variations on the \textbf{\textit{Base Example}} and on all the variations obtained so far.
\end{enumerate}

Table \ref{tab:example_variations} provides some variations for the example in Table \ref{tab:og_examples}. Note that two seperate examples were created through the \textit{'Add irrelevant information'} variation. The first by applying the variation on the \textbf{\textit{Original Example}} and the second by applying it on a previously created example (as directed in Step-4).

To make sure that different \textit{workers} following our protocol make similar types of variations, we hold a trial where each worker created variations from the same 5 seed examples. We observed that barring minor linguistic differences, most of the created examples were the same, thereby indicating the effectiveness of our protocol.

\begin{table*}[h]
	\scriptsize{\centering
		\begin{tabular}{m{10em}m{45em}}
			\toprule
			\textbf{Excerpt of Example} & Beth has 4 packs of red crayons and 2 packs of green crayons. Each pack has 10 crayons in it. \\
			\textbf{Template Form} &	$\big[NAME1\big]$ has $\big[NUM1\big]$ packs of $\big[MOD1\big]$ $\big[OBJp1\big]$ and $\big[NUM2\big]$ packs of $\big[MOD2\big]$ $\big[OBJp1\big]$. \\
			\midrule
			\textbf{Excerpt of Example} & In a game, Frank defeated 6 enemies. Each enemy earned him 9 points.\\
			\textbf{Template Form} & In a game $\big[NAME1\big]$ defeated $\big[NUM1\big]$ $\big[OBJp1\big]$. Each $\big[OBJs1\big]$ earned him $\big[NUM2\big]$ points.\\
			\bottomrule
		\end{tabular}
		\caption{\label{tab:obj_examples} Example uses of tags. Note that in the first example, the word \textit{'packs'} was not replaced since it is a collective. In the second example, the word \textit{'points'} was not replaced because it is too instance-specific and no other word can be used in that place.}
	}
\end{table*}

\begin{table*}[h]
	\scriptsize{\centering
		\begin{tabular}{m{11em}m{50em}}
			\toprule
			\textbf{Seed Example Body} & Beth has 4 packs of crayons. Each pack has 10 crayons in it. She also has 6 extra crayons. \\
			\textbf{Seed Example Question} &	How many crayons does Beth have altogether?\\
			\textbf{Seed Example Equation} & 4*10+6 \\
			\textbf{Script Example Body} & $\big[NAME1\big]$ has $\big[NUM1\big]$ packs of crayons . Each pack has $\big[NUM2\big]$ crayons in it . She also has $\big[NUM3\big]$ extra crayons .\\
			\textbf{Script Example Question} & How many crayons does $\big[NAME1\big]$ have altogether ?\\
			\textbf{Script Example Equation} & $\big[NUM1\big]*\big[NUM2\big]+\big[NUM3\big]$\\
			\bottomrule
		\end{tabular}
		\caption{\label{tab:template_examples} An example of suggested templates. Note that the preprocessing script could not succesfully tag \textit{crayons} as $\big[OBJp1\big]$.}
	}
\end{table*}

\begin{table*}[h]
	\scriptsize{\centering
		\begin{tabular}{m{11em}m{50em}}
			\toprule
			\textbf{Script Example Body} & $\big[NAME1\big]$ has $\big[NUM1\big]$ packs of crayons . Each pack has $\big[NUM2\big]$ crayons in it . She also has $\big[NUM3\big]$ extra crayons .\\
			\textbf{Script Example Question} & How many crayons does $\big[NAME1\big]$ have altogether ?\\
			\midrule
			\textbf{Base Example Body} & $\big[NAME1\big]$ has $\big[NUM1\big]$ packs of \textcolor{green!45!blue}{$\big[OBJp1\big]$} . Each pack has $\big[NUM2\big]$ \textcolor{green!45!blue}{$\big[OBJp1\big]$} in it . She also has $\big[NUM3\big]$ extra \textcolor{green!45!blue}{$\big[OBJp1\big]$} .\\
			\textbf{Base Example Question} & How many \textcolor{green!45!blue}{$\big[OBJp1\big]$} does $\big[NAME1\big]$ have altogether ?\\
			\midrule
			\textbf{$\big[OBJ1\big]$} & crayon, crayons\\
			\textbf{Alternate for $\big[OBJ1\big]$} & pencil, pencils\\
			\textbf{Alternate for $\big[MOD\big]$} & small, large\\
			\bottomrule
		\end{tabular}
		\caption{\label{tab:og_examples} An example of editing the Suggested Templates. The edits are indicated in \textcolor{green!45!blue}{green}.}
	}
\end{table*}

\begin{table*}[h]
	\scriptsize {\centering
		\begin{tabular}{m{10em}m{51em}}
			\toprule
			\textbf{\textit{Base Example Body}} & $\big[NAME1\big]$ has $\big[NUM1\big]$ packs of $\big[OBJp1\big]$. Each pack has $\big[NUM2\big]$ $\big[OBJp1\big]$ in it. She also has $\big[NUM3\big]$ extra $\big[OBJp1\big]$\\
			\textbf{\textit{Base Example Question}} & How many $\big[OBJp1\big]$ does $\big[NAME1\big]$ have altogether ?\\
			\textbf{\textit{Base Example Equation}} & $\big[NUM1\big]*\big[NUM2\big]+\big[NUM3\big]$\\
			\midrule
			\midrule
			\textbf{Category} & Question Sensitivity\\
			\textbf{Variation} & Same Object, Different Structure\\
			\midrule
			\textbf{Variation Body} & $\big[NAME1\big]$ has $\big[NUM1\big]$ packs of $\big[OBJp1\big]$. Each pack has $\big[NUM2\big]$ $\big[OBJp1\big]$ in it. She also has $\big[NUM3\big]$ extra $\big[OBJp1\big]$. \\
			\textbf{Variation Question} & How many $\big[OBJp1\big]$ does $\big[NAME1\big]$ have \textcolor{red!65!blue}{in packs?}\\ \textbf{Variation Equation} & $\big[NUM1\big]*\big[NUM2\big]$\\
			\midrule
			\midrule
			\textbf{Category} & Structural Invariance\\
			\textbf{Variation} & Add irrelevant information\\
			\midrule
			\textbf{Variation Body} & $\big[NAME1\big]$ has $\big[NUM1\big]$ packs of $\big[OBJp1\big]$ \textcolor{green!45!blue}{and $\big[NUM4\big]$ packs of $\big[OBJp2\big]$}. Each pack has $\big[NUM2\big]$ $\big[OBJp1\big]$ in it. She also has $\big[NUM3\big]$ extra $\big[OBJp1\big]$.  \\
			\textbf{Variation Question} & How many $\big[OBJp1\big]$ does $\big[NAME1\big]$ have altogether ?\\ \textbf{Variation Equation} & $\big[NUM1\big]*\big[NUM2\big]+\big[NUM3\big]$\\
			\midrule
			\textbf{Variation Body} & $\big[NAME1\big]$ has $\big[NUM1\big]$ packs of $\big[OBJp1\big]$ \textcolor{green!45!blue}{and $\big[NUM4\big]$ packs of $\big[OBJp2\big]$}. Each pack has $\big[NUM2\big]$ $\big[OBJp1\big]$ in it. She also has $\big[NUM3\big]$ extra $\big[OBJp1\big]$.  \\
			\textbf{Variation Question} & How many $\big[OBJp1\big]$ does $\big[NAME1\big]$ have \textcolor{red!65!blue}{in packs?}\\ \textbf{Variation Equation} & $\big[NUM1\big]*\big[NUM2\big]$\\
			\bottomrule
		\end{tabular}
		
		\caption{\label{tab:example_variations} Example Variations}
	}
\end{table*}
\section{Analyzing Attention Weights}

In Table \ref{tab:attn_wts_sup}, we provide more examples to illustrate the specific word to equation correlation that the constrained model learns.

\begin{table*}[t]
	\small{\centering
		\begin{tabular}{p{35em}P{10em}P{3em}}
			\toprule
			\textbf{Input Problem} & \textbf{Predicted Equation} & \textbf{Answer} \\
			\midrule
			Mike had 8 games. After he gave some to his friend he had 5 \colorbox{caribbeangreen}{left}. How many games did he give to his friend? & 8 - 5 & 3 \cmark \\
			After Mike gave some games to his friend he had 5 \colorbox{caribbeangreen}{left}. If he had 8 games initially, how many games did he give to his friend? & 5 - 8 & -3 \xmark \\
			\midrule
			Jack bought 5 radios but \colorbox{caribbeangreen}{only} 2 of them worked. How many radios did not work? & 5 - 2 & 3 \cmark \\
			Jack bought 5 radios but \colorbox{caribbeangreen}{only} 2 of them worked. How many more radios did not work than those that did? & 5 - 2 & 3 \xmark \\
			\midrule
			Ross had 6 marbles. He \colorbox{caribbeangreen}{sold} 2 marbles to Joey. How many marbles does Ross have now? & 6 - 2 & 4 \cmark \\
			Ross had 6 marbles. Joey \colorbox{caribbeangreen}{sold} 2 marbles to Ross. How many marbles does Ross have now? & 6 - 2 & 4 \xmark \\
			\midrule
			Bob \colorbox{caribbeangreen}{collected} 7 cans. He lost 3 of them. How many cans does Bob have now? & 7 - 3 & 4 \cmark \\
			Bob had 7 cans. He \colorbox{caribbeangreen}{collected} 3 more. How many cans does Bob have now? & 7 - 3 & 4 \xmark \\
			\midrule
			Joey had 9 pens. he \colorbox{caribbeangreen}{used} 4 of them. How many pens does he have now? & 9 - 4 & 5 \cmark \\
			Joey \colorbox{caribbeangreen}{used} 4 pens. If he had 9 pens intially, how many pens does he have now? & 4 - 9 & -5 \xmark \\
			\midrule
			Jill read 30 pages in 10 days. How many pages did she read \colorbox{caribbeangreen}{per} day? & 30 / 10 & 3 \cmark \\
			Jill can read 3 pages \colorbox{caribbeangreen}{per} day. How many pages can she read in 10 days? & 3 / 10 & 0.33 \xmark \\
			\midrule
			Mary's hair was 15 inches long. After she did a haircut, it was 10 inches long . how much did she cut \colorbox{caribbeangreen}{off} ? & 15 - 10 & 5 \cmark \\
			Mary cut \colorbox{caribbeangreen}{off} 5 inches of her hair. If her hair is now 10 inches long, how long was it earlier? & 5 - 10 & -5 \xmark \\
			\bottomrule
		\end{tabular}
		\caption{\label{tab:attn_wts_sup} Attention paid to specific words by the constrained model.}
	}
\end{table*}
\section{Examples of Simple Problems}

In Table \ref{tab:failure_examples}, we provide a few simple examples from \dataname{} that the best performing Graph2Tree model could not solve.

\begin{table*}[t]
	\small{\centering
		\begin{tabular}{p{30em}P{8em}P{9em}}
			\toprule
			\textbf{Input Problem} & \textbf{Correct Equation} & \textbf{Predicted Equation} \\
			\midrule
			Every day ryan spends 6 hours on learning english and 2 hours on learning chinese. How many more hours does he spend on learning english than he does on learning chinese? & 6 - 2 & 2 - 6 \\
			\midrule
			In a school there are 34 girls and 841 boys. How many more boys than girls does the school have? & 841 - 34 & 34 - 841 \\
			\midrule
			David did 44 push-ups in gym class today. David did 9 more push-ups than zachary. How many push-ups did zachary do? & 44 - 9 & 44 + 9 \\
			\midrule
			Dan has \$ 3 left with him after he bought a candy bar for \$ 2. How much money did he have initially? & 3 + 2 & 3 - 2 \\
			\midrule
			Jake has 11 fewer peaches than steven. If jake has 17 peaches. How many peaches does steven have? & 11 + 17 & 17 - 11 \\
			\midrule
			Kelly gives away 91 nintendo games. How many did she have initially if she still has 92 games left? & 91 + 92 & 92 - 91 \\
			\midrule
			Emily is making bead necklaces for her friends. She was able to make 18 necklaces and she had 6 beads. How many beads did each necklace need? & 18 / 6 & 6 / 18 \\
			\midrule
			Frank was reading through some books. Each book had 249 pages and it took frank 3 days to finish each book. How many pages did he read per day? & 249 / 3 & ( 249 * 3 ) / 3 \\
			\midrule
			A mailman has to give 5 pieces of junk mail to each block. If he gives 25 mails to each house in a block, how many houses are there in a block? & 25 / 5 & 5 / 25 \\
			\midrule
			Faye was placing her pencils and crayons into 19 rows with 4 pencils and 27 crayons in each row. How many pencils does she have? & 19 * 4 & 19 * 27 \\
			\midrule
			White t - shirts can be purchased in packages of 53. If mom buys 57 packages of white t - shirts and 34 trousers, How many white t - shirts will she have? & 53 * 57 & ( 53 * 57 ) + 34 \\
			\midrule
			An industrial machine can make 6 shirts a minute. It worked for 5 minutes yesterday and for 12 minutes today. How many shirts did machine make today?  & 6 * 12 & 5 + 12 \\
			\bottomrule
		\end{tabular}
		\caption{\label{tab:failure_examples} Some simple examples from \dataname{} on which the best performing Graph2Tree model fails.}
	}
\end{table*}
\section{Ethical Considerations}

In this paper, we consider the task of automatically solving Math Word Problems (MWPs). Our work encourages the development of better systems that can robustly solve MWPs. Such systems can be deployed for use in the education domain. E.g., an application can be developed that takes MWPs as input and provides detailed explanations to solve them. Such applications can aide elementary school students in learning and practicing math.

We present a challenge set called \dataname{} of one-unknown English Math Word Problems. \dataname{} is created in-house by the authors themselves by applying some simple variations to examples from ASDiv-A \cite{miao-etal-2020-diverse}, which is a publicly available dataset. We provide a detailed creation protocol in Section \ref{asec:protocol}. We are not aware of any risks associated with our proposed dataset.

To provide an estimate of the energy requirements of our experiments, we provide the details such as computing platform and running time in Section \ref{implementation}. Also, in order to reduce carbon costs from our experiments, we first perform a broad hyperparameter search over only a single fold for the datasets and then run the cross validation experiment over a select few hyperparameters.

\end{document}